\documentclass[letterpaper, 10pt, conference]{ieeeconf}

\IEEEoverridecommandlockouts                             
\usepackage[top=0.72in, bottom=0.72in, left=0.72in, right=0.72in]{geometry}

\usepackage{graphicx}
\usepackage{amsmath} 
\usepackage{amssymb}

\usepackage{amsthm}
\theoremstyle{definition}
\newtheorem{definition}{Definition}
\usepackage{booktabs}
\usepackage{hyperref}
\hypersetup{
    colorlinks=true,
    linkcolor=blue,
    filecolor=magenta,      
    urlcolor=blue,
}
\usepackage{xcolor}
\usepackage{multirow}
\let\labelindent\relax
\usepackage{enumitem}
\usepackage{graphicx}
\usepackage{subcaption}
\usepackage{float}
\usepackage{algorithm}
\usepackage[end]{algpseudocode}
\usepackage{booktabs}
\usepackage{array}
\usepackage{caption}

\setlist[itemize]{nosep, leftmargin=*}
\setlist[enumerate]{nosep, leftmargin=*}
\setlist{nosep}
\allowdisplaybreaks
\usepackage{titlesec}
\titlespacing{\section}{0pt}{*0.6}{*0.75}
\titlespacing{\subsection}{0pt}{*0.6}{*0.6}
\titlespacing{\subsubsection}{0pt}{*0.5}{*0.4}
\usepackage{microtype}
\makeatletter
\def\tagform@#1{\maketag@@@{\ignorespaces(#1)\unskip\@@italiccorr}}
\makeatother

\usepackage{lettrine}

\newcommand{\IEEEPARstart}[2]{%
\lettrine[
    lines=2,
    lhang=0,
    loversize=0.15,
    findent=0pt,
    nindent=0pt
]{#1}{#2}%
}

\usepackage{etoolbox}
\makeatletter
\def\@maketitle{%
  \newpage
  \null
  \vskip 2em
  \begin{center}
    \vspace*{-2.8em}
    {\normalfont\footnotesize
    \textcolor{gray}{This paper has been accepted for publication at the 2026 IEEE/RSJ International Conference on Intelligent Robots and Systems}
    \textcolor{blue}{\href{https://2026.ieee-iros.org/}{(IROS 2026)}}\par}
    \vspace{2.2em}
    {\LARGE \bf \@title \par}
    \vskip 1.5em
    {\rm \lineskip .5em
    \begin{tabular}[t]{c}\@author
    \end{tabular}\par}
  \end{center}
  \par
  \vskip 1.5em}
\makeatother

\title{\LARGE \bf
Safe Overtaking for Autonomous Racing Using Hierarchical Optimization and Learning-Based Control
}

\author{Hassan Jardali, Kai Yin, Lantao Liu %
\thanks{Hassan Jardali and Lantao Liu are with Luddy School of Informatics, Computing, and Engineering, Indiana University (emails: hjardali@iu.edu, lantao@iu.edu).\newline
\indent Kai Yin is with the Machine Learning Science, Expedia Group, Austin, TX 78758 USA
 (email: yinkai1000@gmail.com).
}}

\begin{document}

\maketitle
\thispagestyle{empty}
\pagestyle{empty}

\vspace*{-0.5in}
\begin{abstract}

Autonomous racing overtaking requires balancing competitive performance with safety under nonlinear vehicle dynamics and real-time constraints. Model Predictive Control (MPC) combined with 
Control Barrier Functions (CBFs) provides a principled mechanism for certifying forward invariance of a safe set. However, commonly used fixed-decay discrete-time CBF formulations can become overly conservative in interactive racing scenarios, limiting overtaking performance and requiring manual tuning across track conditions.
This paper proposes a hierarchical overtaking framework that explicitly separates maneuver-level decision making from safety-certified trajectory control, reducing conservatism while preserving safety. A high-level Mixed-Integer Quadratic Program (MIQP) resolves the combinatorial passing-side selection problem by selecting a feasible overtaking topology, while a nonlinear Frenet-frame MPC enforces vehicle dynamics and safety through embedded discrete-time CBF constraints. 
This decomposition isolates the combinatorial complexity of maneuver selection from the continuous trajectory optimization.
To further mitigate the sensitivity of fixed-decay barrier constraints, a reinforcement learning policy adapts the discrete-time CBF decay parameter online, enabling context-dependent modulation of safety margins without directly controlling vehicle inputs. Simulation and scaled-hardware experiments show that no single fixed decay parameter achieves uniformly strong performance across tracks, whereas the adaptive strategy attains the highest aggregate success rate and consistently strong safety–performance trade-offs without per-track tuning, improving robustness to environment variation while maintaining safety constraint satisfaction in nominal operation.

\end{abstract}

\section{INTRODUCTION}
\IEEEPARstart{A}{utonomous} racing has emerged as a premier testbed for pushing the boundaries of robotic planning and control algorithms at the limits of vehicle handling \cite{indyautonomouschallengeIndyAutonomous, a2rlAutonomousRace}. Unlike standard autonomous driving, which prioritizes passenger comfort and strict adherence to traffic rules, racing demands that agents operate at the edge of the friction circle to minimize lap times.

Overtaking represents a particularly demanding maneuver, requiring a careful balance between two often conflicting objectives: competitiveness and safety. A successful overtake necessitates operating near the vehicle’s dynamic limits while guaranteeing collision avoidance at all times. 
Excessively conservative strategies for safety preservation often sacrifice track position, whereas insufficiently cautious behavior can lead to failures.

\begin{figure}[htbp]
    \centering
    \includegraphics[ height=3cm, trim=200 50 150 100, clip]{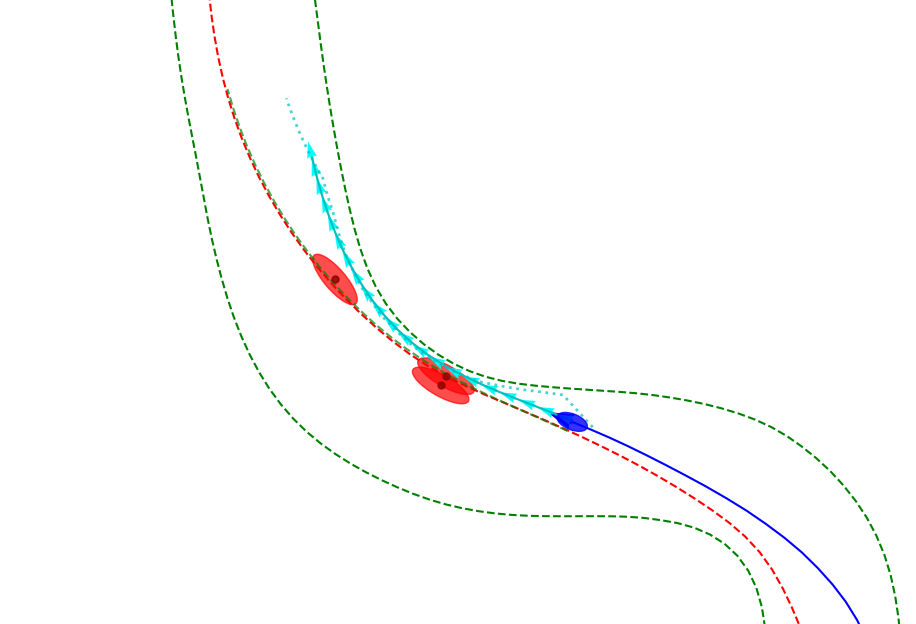}
    \hfill
    \includegraphics[height=3cm, trim=0 0 0 0, clip]{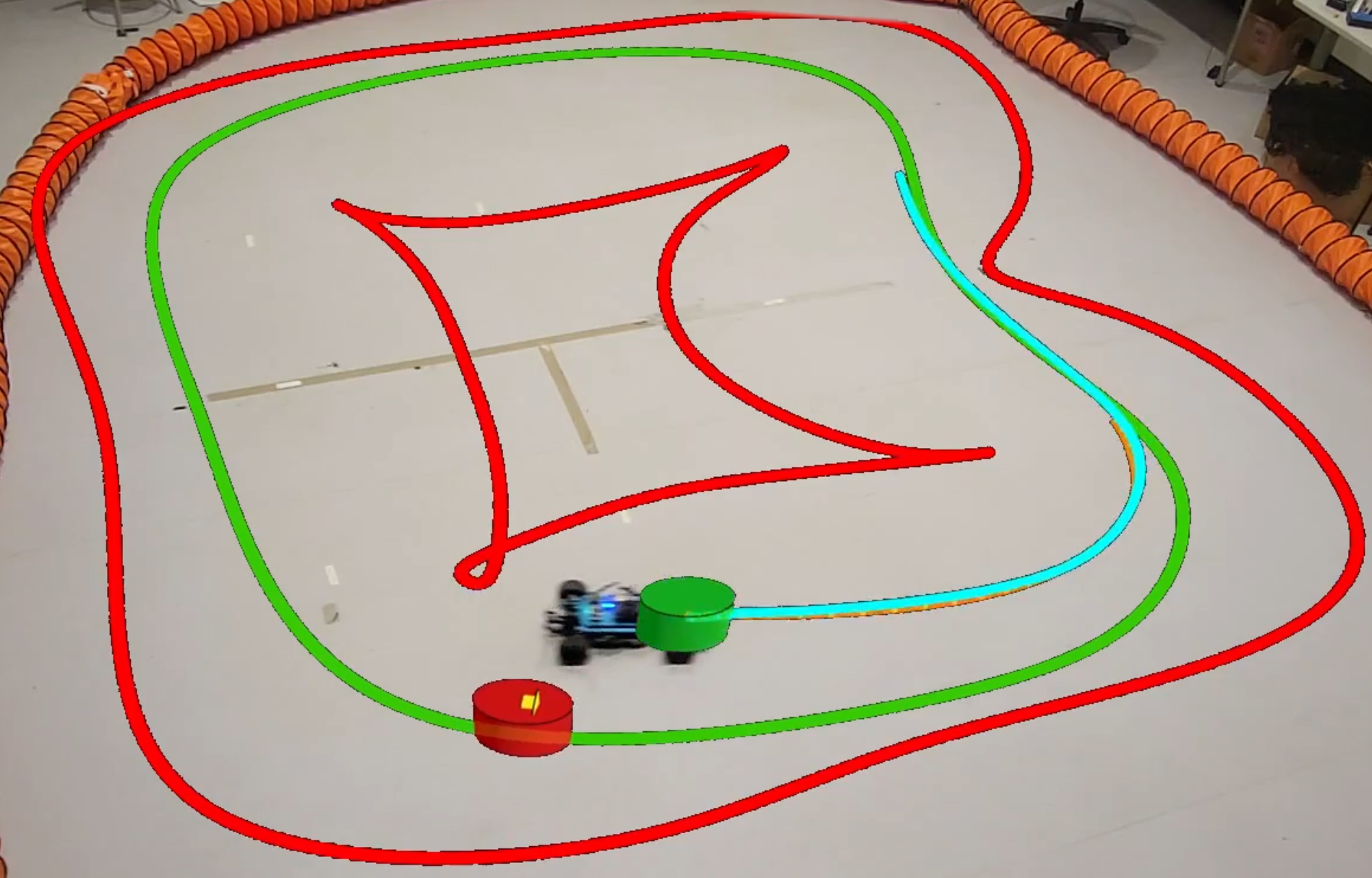}
    \caption{\small Snapshots of the proposed framework. Left: simulation on the Laguna Seca track. Right: scaled-vehicle hardware demonstration with a virtual (ghost) opponent. The ego vehicle is shown in blue and opponent vehicles in red. The target raceline is depicted in red. The high-level MIQP planner output is illustrated as a dotted light-blue trajectory, and the solid blue curve represents the MPC–CBF predicted trajectory.}
    \label{fig:snapshots}
\end{figure}

From a control-theoretic perspective, safety can be formalized as the forward invariance of a certified safe set. Control Barrier Functions (CBFs)~\cite{ames2019control} provide a principled mechanism for enforcing this invariance within optimization-based controllers by imposing state-dependent constraints that guarantee safety over time. When integrated into Model Predictive Control (MPC), CBF constraints enable safety filtering while respecting vehicle dynamics and actuator limits already encoded within the MPC formulation. 
In practice, many implementations adopt \emph{fixed-decay} \emph{discrete-time CBFs}~\cite{9483029}, 
where safety is enforced through a one-step contraction condition governed by a constant decay parameter. Intuitively, this decay parameter dictates how aggressively the vehicle is permitted to approach the boundary of the safe set, acting as a tunable buffer for collision avoidance.
While computationally convenient, 
this formulation enforces safety locally at each time step rather than explicitly \textit{reasoning over maneuver-level decisions across a planning horizon}.
Consequently, fixed-decay discrete-time CBF constraints may introduce overly conservative behavior in competitive overtaking scenarios, where dynamically recoverable states 
can be prematurely excluded due to rigid decay parameter selection and the lack of explicit \textit{long-horizon maneuver awareness}.

To address these limitations, we propose a hierarchical framework that balances three core requirements of multi-vehicle racing: \textbf{safety}, \textbf{dynamic feasibility}, and \textbf{competitive performance}.
A Mixed-Integer Quadratic Program (MIQP)~\cite{wolsey2007mixed} handles high-level discrete overtaking topology selection to achieve competitive
performance, while dynamic feasibility is managed by a nonlinear MPC in the Frenet frame, allowing operation near tire friction limits. Safety is enforced through discrete-time CBF constraints embedded within the MPC, certifying forward invariance of a velocity-aware collision-avoidance set under nominal conditions. Finally, to further maintain competitive performance, a Reinforcement Learning (RL) policy adapts the CBF decay parameter online. This adaptation enables context-dependent modulation of safety margins, mitigating conservatism during real-time racing maneuvers. 
The main contributions of this work include:

\begin{itemize}
    \item A unified hierarchical architecture combines high-level mixed-integer quadratic programming (MIQP) for discrete overtaking topology selection with low-level MPC–CBF control. The MIQP resolves homotopy decisions (e.g., passing side) to handle non-convexity, 
    while the MPC–CBF layer maintains dynamic feasibility and provides formal safety certification through discrete-time CBF constraints.
    
    \item An RL-based adaptation mechanism for the discrete-time CBF decay parameter, introducing context-dependent safety modulation that reduces conservatism while preserving the structure of the safety constraints.
    
    \item Simulation results show that adaptive decay achieves the highest aggregate success rate across multiple tracks, improving robustness without manual tuning. Scaled-hardware experiments confirm real-time deployability of the hierarchical framework. The implementation is released as open source to support reproducibility at: \href{https://github.com/hassanjardali/SOfAR}{github.com/hassanjardali/SOfAR}.
\end{itemize}

\section{Related Work}

\subsection{Overtaking and Trajectory Planning Frameworks}
Overtaking in autonomous racing has been addressed through hierarchical, optimization-based, and graph-search planning frameworks. Optimization-based approaches, including nonlinear MPC formulations, embed opponent models directly within the prediction dynamics to derive overtaking strategies implicitly from the optimal control solution \cite{8206086}. While conceptually unified, such formulations can impose substantial computational demands in multi-agent scenarios.

Many methods adopt hierarchical decompositions to improve real-time tractability. 
High-level planners, such as dynamic programming \cite{Liniger2015AutonomousRacing, toschi2025modular} or spatio-temporal visibility graphs \cite{10422566}, select feasible passing corridors or maneuver envelopes, which are then enforced as constraints within a lower-level MPC or time-optimal controller. Similarly, multilayer graph-based planners generate discrete trajectory candidates that enable behavioral selection in dynamic racing environments \cite{stahl2019multilayer}. Sampling-based and Frenet-frame trajectory generation methods \cite{werling2010optimal, raji2022motion} further support reactive collision avoidance. 


Despite these diversity of planning strategies, overtaking fundamentally involves discrete maneuver decisions such as selecting the relative passing side or corridor that partition the feasible motion space into distinct topological classes. While many frameworks encode these choices implicitly through corridor selection or trajectory sampling, the underlying combinatorial structure remains an essential component of the problem\cite{Liniger2015AutonomousRacing}.

\subsection{Hierarchical Planning and Mixed-Integer Formulations}

Explicitly modeling this combinatorial structure motivates Mixed-Integer Programming (MIP) formulation~\cite{dollar2021multilane, miller2018efficient}, where discrete maneuver decisions are represented using binary variables.
In particular, Mixed-Integer Quardratic Programming (MIQP) enables the integration of discrete decision-making with continuous trajectory optimization within a unified optimal control framework~\cite{ioan2021mixed}. However, the computational complexity of mixed-integer formulations grows rapidly with increasing prediction horizons, limiting their direct application in real-time settings~\cite{pia2017mixed}.

To mitigate this challenge, many approaches adopt hierarchical planning architectures that decompose the problem into discrete and continuous layers~\cite{gratzer2024two}. In these frameworks, a high-level planner, often implemented as an MIQP, selects a feasible homotopy class or maneuver by resolving discrete interaction decisions among vehicles~\cite{park2015homotopy}. By isolating combinatorial reasoning from low-level trajectory optimization, the overall computational burden is significantly reduced while preserving structured decision-making capabilities~\cite{quirynen2024real}. A downstream continuous controller subsequently refines and tracks the selected maneuver, ensuring dynamic feasibility, constraint satisfaction, and real-time responsiveness in multi-agent environments. 
Next we will review a widely established safety-certified control mechanism. 

\subsection{Adaptive Safety Certification and Control Barrier Functions}
Control Barrier Functions (CBFs) have emerged as a principled mechanism for enforcing forward invariance of safety sets within optimization-based controllers, including MPC formulations \cite{9483029}. In autonomous racing, CBF constraints have been integrated with trajectory optimization to guarantee collision avoidance while preserving time-optimal behavior \cite{9811969}. More recently, learning-based approaches have been explored to mitigate the conservatism introduced by fixed safety margins and manually tuned controller parameters. Reinforcement learning (RL) has been used as a high-level selector to adapt planning cost functions in agile driving scenarios \cite{langmann2025reinforcement}, enabling context-dependent modulation of aggressive and conservative behaviors. In the context of safety-critical control, RL has also been employed to tune objective weights or CBF-related parameters within MPC-CBF architectures, typically through parameterized controllers and bilevel optimization schemes \cite{sabouni2024reinforcement}. These methods aim to balance performance and safety while retaining the forward-invariance structure of CBF constraints. Concurrent efforts investigate learning CBF representations or non-greedy safety filters to improve performance while maintaining formal guarantees \cite{wijayatunga2026learning}.

Despite these advances, most existing approaches either adapt cost functions while keeping safety constraints fixed, or tune multiple controller parameters simultaneously, making it difficult to isolate the impact of safety-margin modulation. In contrast, structured adaptation of a single, interpretable discrete-time CBF decay parameter offers a direct mechanism for modulating conservatism while preserving the underlying safety constraint formulation.

\section{Problem Formulation and Preliminaries}\label{sec:preliminaries}



We begin by introducing the notation that will be used throughout this paper. Let $\mathbb{R}$ denote the set of real numbers and $\mathbb{R}_+$ the set of non-negative real numbers. We consider a discrete-time framework where time instances are indexed by $k \in \mathbb{N}$, corresponding to continuous time $t = k T_s$ with a sampling period $T_s > 0$. 
The state of the ego vehicle is denoted by $x_k \in \mathcal{X} \subseteq \mathbb{R}^n$, and the control input by $u_k \in \mathcal{U} \subseteq \mathbb{R}^m$, where $\mathcal{X}$ and $\mathcal{U}$ represent the admissible state and control input spaces, respectively. For a continuously differentiable scalar function $h: \mathcal{X} \rightarrow \mathbb{R}$, the 
set $\mathcal{S}$ is defined as the zero superlevel set:
    $\mathcal{S} = \{ x \in \mathcal{X} \mid h(x) \ge 0 \}.$
Throughout this work, $\|\cdot\|$ denotes the standard Euclidean $L_2$ norm.

\subsection{Discrete-Time Vehicle Dynamics Model}


We model the ego vehicle as a discrete-time control-affine nonlinear system
\begin{equation}
    x_{k+1} = f(x_k) + g(x_k)u_k,
\end{equation}
where the functions
$f:\mathcal{X}\rightarrow \mathbb{R}^n$ and
$g:\mathcal{X}\rightarrow \mathbb{R}^{n\times m}$
represent the drift dynamics and input matrix, respectively,
and are assumed to be smooth. 
This control-affine structure explicitly separates the uncontrolled
system evolution from the control-dependent component and enables
the following construction of control barrier function (CBF) constraints that
are affine in the control input.
The vehicle motion is represented in the Frenet frame relative to the
track centerline, which decouples longitudinal progress from lateral
deviation and is well suited for racing scenarios on structured tracks.

In the Frenet frame, the state vector is defined as
$x_k = [s_k,\; \eta_k,\; e_k,\; v_k,\; \delta_k]^\top$,
where $s_k$ denotes longitudinal progress along the reference path,
$\eta_k$ the lateral deviation from the centerline,
$e_k$ the heading error relative to the path tangent,
$v_k$ the longitudinal velocity, and
$\delta_k$ the steering angle.
The control input is given by
$
u_k = [a_k,\; \dot{\delta}_k]^\top,
$
where $a_k$ is longitudinal acceleration and
$\dot{\delta}_k$ is steering rate.

The discrete-time dynamics are obtained by numerically integrating a
continuous-time kinematic bicycle model over the sampling interval $T_s$
using a fourth-order Runge--Kutta scheme. The continuous-time
bicycle model equations are standard and provided in the Appendix.
This modeling choice captures the dominant vehicle kinematics while
remaining computationally tractable for optimization-based control.
In practice, the vehicle dynamics may be fully nonlinear. The control-affine representation via bicycle model can be obtained via local linearization of the discretized dynamics within each MPC update, which maintains sufficient modeling precision for real-time optimization.

\subsection{Safe Set Definition}

In the context of autonomous overtaking, safety-critical collision avoidance is formulated through a certified safe set defined as the intersection of superlevel sets associated with each opponent vehicle:
\begin{equation}
\mathcal{S} = \bigcap_{j=1}^{N_o} \mathcal{S}_j, 
\quad \text{where } 
\mathcal{S}_j = \{ x \in \mathcal{X} \mid h_j(x) \ge 0 \},
\end{equation}
with continuously differentiable 
functions
$h_j : \mathcal{X} \rightarrow \mathbb{R}$. 
For notational simplicity, the discrete-time index is omitted.

\textbf{Velocity-aware collision avoidance constraints:}
For each opponent vehicle $j$, an elliptical safety function is defined as
\begin{equation}
h_j(x) = 
\left( \frac{s - s_j}{a_j(v)} \right)^2 +
\left( \frac{\eta - \eta_j}{b} \right)^2 - 1,
\end{equation}
where $(s_j, \eta_j)$ denotes the predicted opponent position in the Frenet frame. 
The parameters $a_j(v)$ and $b$ define the longitudinal and lateral semi-axes of an elliptical safety boundary centered at the opponent vehicle. 
The lateral semi-axis $b$ specifies a fixed lateral safety margin, while the longitudinal semi-axis $a_j(v)$ scales with relative velocity according to
$a_j(v) = a_0 + \frac{v_{\text{rel}}^2}{2 a_{\min}}$,
where $v_{\text{rel}} = \max(0, v - v_j)$. 
The parameter $a_{\min}$ represents the maximum available braking deceleration and captures the longitudinal stopping capability of the ego vehicle under the adopted vehicle dynamics model.
This velocity-aware formulation enlarges the longitudinal safety margin at higher closing speeds according to a braking-distance model.

\medskip
Track boundary constraints are enforced separately as state constraints within the MPC formulation. Specifically,
\begin{equation}
\eta_{\min,k} \le \eta_k \le \eta_{\max,k},
\end{equation}
where $\eta_{\max,k}$ and $\eta_{\min,k}$ denote the outer and inner track boundaries at prediction step $k$ along the control horizon.

\subsection{Discrete-Time Control Barrier Functions}
Forward invariance is a fundamental concept for certifying safety. 
A set $\mathcal{S}$ is said to be \textit{forward invariant} under the control policy $u_k = \pi(x_k)$ if $x_0 \in \mathcal{S}$ implies $x_k \in \mathcal{S}$ for all $k \ge 0$.
For discrete-time systems, forward invariance of $\mathcal{S} = \{x \vert h(x) \ge 0\}$ can be imposed by the one-step condition
\begin{equation}
h(x_{k+1}) \ge (1 - \gamma) h(x_k),
\end{equation}
where $\gamma \in (0,1]$ is a decay parameter that regulates the allowable decrease of the safety function. 
Smaller values of $\gamma$ restrict the rate at which $h(x_k)$ may decrease, resulting in more conservative behavior and larger safety margins. 
Larger values of $\gamma$ relax this restriction, permitting more aggressive trajectories at the cost of reduced safety margins.


\begin{definition}[Discrete-Time Control Barrier Function]
A function $h : \mathcal{X} \rightarrow \mathbb{R}$ is a discrete-time Control Barrier Function (DCBF) if there exists $\gamma \in (0,1]$ such that for all $x_k \in \mathcal{S}$, there exists a control input $u_k \in \mathcal{U}$ satisfying
\begin{equation}
h(f(x_k)+ g(x_k)u_k) \ge (1 - \gamma) h(x_k).
\end{equation}
\end{definition}

When embedded within an MPC framework, this condition is imposed at each prediction step to certify forward invariance of the collision-avoidance set when the barrier constraints are satisfied. 
In standard DCBF–MPC implementations, $\gamma$ is selected as a fixed design parameter, introducing a trade-off between safety conservatism and overtaking performance. 
In racing scenarios, this fixed-decay selection can be overly restrictive, motivating adaptive modulation of $\gamma$.

\section{Hierarchical Planning and Control Framework}



\begin{figure}[!htbp]
    \centering
    \includegraphics[width=1.0\linewidth, trim=180 250 180 250, clip]{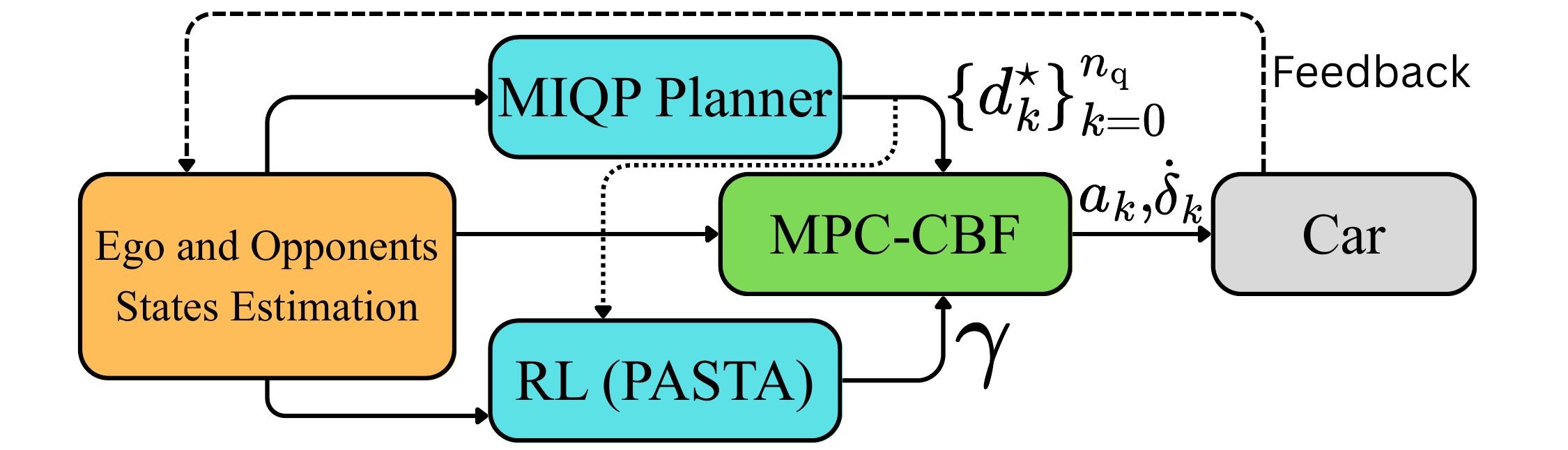}
    \caption{\small Overview of hierarchical planning and control architecture.}
    \label{fig:sys_diagram}
\end{figure}

This section presents a hierarchical planning and control framework for real-time execution as shown in Fig.~\ref{fig:sys_diagram}. The architecture separates discrete maneuver selection from continuous trajectory optimization. 
A high-level planner resolves combinatorial overtaking decisions, while a low-level MPC–CBF controller generates safe control inputs. At each control step, ego and opponent states are estimated, transformed into the Frenet frame, and propagated using a constant-velocity model. An RL policy adapts the CBF decay parameter online, enabling context-dependent safety modulation without altering the barrier formulation.

\subsection{High-Level Overtaking Decision via MIQP} 
Overtaking inherently involves discrete maneuver decisions, such as choosing whether to pass an opponent on the left or on the right. These mutually exclusive choices induce distinct feasible motion patterns that cannot be represented by a single convex set of trajectories \cite{Liniger2015AutonomousRacing}. 
When incorporated into a nonlinear MPC formulation, collision-avoidance constraints must account for multiple alternative maneuver possibilities, resulting in a non-convex feasible region composed of several disconnected subsets. The decision maker must therefore reason over competing trajectory candidates, giving rise to a fundamentally combinatorial decision structure that can lead to local minima, oscillatory behavior, or infeasibility under real-time constraints.

To explicitly address this combinatorial structure, we employ an MIQP to determine the overtaking decision prior to continuous trajectory optimization. 
The high-level planner solves an MIQP in the Frenet frame to generate a lateral reference trajectory over a coarse horizon. The optimization selects the overtaking maneuver (e.g., pass left or right) and produces a smooth lateral deviation profile relative to the nominal racing line, while longitudinal progress is treated as given. This separation enables the system to plan discrete maneuvers at a slower update rate while executing continuous control commands in real time.

Let $n_q$ denote the planning horizon length, $k\in [n_q]\triangleq\{1,\dotsm n_q\}$ be the planning step, and $i \in [I]\triangleq\{1,\dots,I\}$ index the
opponent vehicles. The high-level overtaking planner is formulated as the
following MIQP:
\begin{subequations}\label{eq:miqp}
\begin{align}
\min_{\{d_k\},\{b_i\}} \;
& \sum_{k=0}^{n_q} \Big(
\| d_k - \eta_{\mathrm{ref},k} \|^2  \notag\\
&\hspace{1.5em}
+ \| d_{k+1} - d_k \|^2
+ \| d_k - d_k^{\mathrm{prev}} \|^2
\Big)
\\
\text{s.t.}\quad 
& \eta_{\min,k} \le d_k \le \eta_{\max,k},
\qquad \forall k\in[n_q], \label{track_bd_constr}\\
& d_k \le n_{\mathrm{obs}}^i - w_{\mathrm{safe}}
    + M b_i + \xi_{k,i},
\qquad \forall k,i, \label{big-m-a}\\
& d_k \ge n_{\mathrm{obs}}^i + w_{\mathrm{safe}}
    - M(1-b_i) - \xi_{k,i},
\, \forall k, i, \label{big-m-b}\\
& b_i \in \{0,1\},
\quad \forall i\in[I],
\end{align}
\end{subequations}
where the decision variable $d_k$ represents the planned lateral deviation
relative to the nominal racing line $\eta_{\mathrm{ref},k}$, while
$d_k^{\mathrm{prev}}$ denotes the solution obtained at the previous
control cycle and serves as a temporal regularization term promoting
inter-iteration consistency.
Constraints~\eqref{track_bd_constr} impose track boundary limits.
Collision avoidance with opponent vehicle $i$ is encoded via the
Big-$M$ constraints~\eqref{big-m-a}--\eqref{big-m-b}, where
$n_{\mathrm{obs}}^i$ denotes the predicted lateral position of opponent
$i$ in the Frenet frame, and $M$ is a
sufficiently large constant used to linearize mutually exclusive
maneuver constraints under different binary decisions $b_i\in \{0,1\}$. The parameter
$w_{\mathrm{safe}}$ specifies a prescribed lateral safety margin, and
$\xi_{k,i}$ is a slack variable introduced to preserve feasibility in
the presence of modeling uncertainty. The binary decision variable
$b_i$ determines the overtaking side, with $b_i=1$ enforcing one side passing and
$b_i=0$ corresponding to passing on opposite side of the opponent.
Binary consistency is enforced across the planning
horizon to prevent oscillatory switching between maneuver decisions.

The mixed-integer constraints in \eqref{eq:miqp} encode safe overtaking or obstacle avoidance as a geometric partition of the feasible lateral motion space.
The feasible set may be written as a union of convex regions $\bigcup_{i}\Omega(\{b_i\})$, where $\Omega(\cdot)$ refers to feasible region depending on the entire set of binary decision variable $\{b_i\}$. Each assignment $\{b_i\}$ defines a convex quadratic program and a distinct homotopy class of trajectories $\{d_i\}$. Geometrically, the MIQP performs combinatorial selection among trajectory manifolds while maintaining the smoothness objective. If $b_i\in\{0, 1\}$ is relaxed to continuous variable $b_i\in [0, 1]$, then the feasible set becomes the convex hull of these regions $\text{conv}(\Omega(\{b_i\}))$, where some of the solutions do not correspond to physically realizable maneuvers. Such relaxation may provide a lower bound on the optimal MIQP objective function.

The MIQP is viewed as a planner and therefore executed at a lower frequency than the controller, and its solution provides a dynamically consistent lateral reference trajectory for the downstream controller.
\vspace{-0.02in}
\subsection{Low-Level Control via MPC--DCBF}

The low-level controller tracks the MIQP-generated reference using a nonlinear MPC formulated in the Frenet frame. The discrete-time vehicle dynamics introduced in Section~\ref{sec:preliminaries} are enforced together with actuator and state bounds.
Let $\{ d_k^*\}_{k=0}^{n_q}$ denote the optimal lateral offset sequence along with $\{ b_i^*\}_{i=0}^{I}$ determining the passing side from the MIQP. This solution defines the selected overtaking homotopy class and serves as the lateral component of the reference trajectory for the downstream controller. 
The reference state for the MPC is constructed as
\begin{equation}
x_{\mathrm{ref},k} =
\begin{bmatrix}
s_k^{\mathrm{ref}},
d_k^*,
e_k^{\mathrm{ref}},
v_k^{\mathrm{ref}},
\delta_k^{\mathrm{ref}}
\end{bmatrix}^\top,
\end{equation}
where $s_k^{\mathrm{ref}}$, $e_k^{\mathrm{ref}}$, $v_k^{\mathrm{ref}}$, and $\delta_k^{\mathrm{ref}}$ are obtained from the nominal racing line or propagated reference trajectory. 
For MPC, we consider the following finite-horizon optimization problem
\begin{subequations}\label{eq:mpc}
\begin{align}
\min_{\{u_k\}}
\sum_{k=0}^{n_c}
&\left(
\| x_k - x_{\mathrm{ref},k} \|_{Q}^2
+ \| u_k \|_{R}^2
\right), \\
s.t. \quad x_{k+1} &= f({x}_k, {u}_k), \\
{u}_k &\in \mathcal{U}, \
{x}_k \in \mathcal{X},
\end{align}
\end{subequations}
where $n_c$ is the length of finite horizon, $Q$ and $R$ are standard weighting matrices~\cite{mayne2014model}. In addition, we consider the following different choices of constraints in Eqn~\eqref{eq:mpc}.

\emph{a) Direct DCBF Constraints:}
The discrete-time control barrier condition defined in Section~\ref{sec:preliminaries} is imposed at each prediction stage of the MPC horizon. For each opponent vehicle $i$, the constraint $ (1-\gamma)\, h_i({x}_k) - h_i({x}_{k+1}) \le 0$
is enforced using the predicted obstacle state and the one-step vehicle dynamics.
In addition to obstacle avoidance enforced via DCBF constraints, track boundary and lateral acceleration limits are imposed as standard state constraints within the same nonlinear program.

\emph{b) Soft DCBF Constraints:}
Since safety constraints may become temporarily incompatible with input and state bounds within a finite prediction horizon, strict enforcement can lead to infeasibility of the nonlinear program. Therefore, all nonlinear constraints, including DCBF constraints, are implemented as soft constraints.
For barrier condition, it means the following
\begin{equation}
(1-\gamma)\, h_i({x}_k) - h_i({x}_{k+1})
\le \varepsilon_{k,i},
\end{equation}
where $\varepsilon_{k,i}\ge 0$ are slack variables. When all slack variables satisfy $\varepsilon_{k,i} = 0$ along the executed trajectory, forward invariance of the collision-avoidance set follows directly from the original condition. 
Activation of slack variables corresponds to temporary relaxation of strict invariance to preserve recursive feasibility. 
The relaxation for all constraints are chosen in a way that prioritizes collision avoidance over comfort-related constraints and activates slack only when the nonlinear program would otherwise become infeasible.



\begin{figure}
    \centering
    \includegraphics[width=1.0\linewidth]{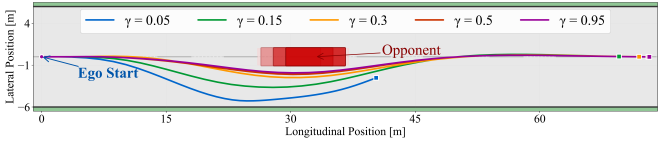}
    \caption{\small Effect of the decay parameter $\gamma$ on the MPC-CBF trajectory evolution over the same temporal window (MIQP disabled).
    }
    \label{fig:gamma_effect}
\end{figure}

\subsection{Adaptive Safety Modulation via Reinforcement Learning}

The discrete-time CBF decay parameter $\gamma \in (0,1]$ regulates the contraction rate of the safety function and directly governs the trade-off between conservative and aggressive behavior. Smaller values of $\gamma$ enforce stricter safety contraction, yielding larger safety margins, whereas larger values permit more aggressive maneuvers at reduced conservatism, as illustrated in Fig.~\ref{fig:gamma_effect}. In competitive racing scenarios, the appropriate safety margin depends on traffic density, relative motion, and local track geometry, rendering fixed decay selection suboptimal.

To enable context-dependent safety modulation, we employ a multi-objective reinforcement learning (MORL) framework to adapt the DCBF parameter $\gamma$ online. Given the environment observation $o$, which captures the states of both ego and opponent vehicles, the objective is to learn a policy $\pi(o)$ that selects an action $a$. In our formulation, the action corresponds to the value of the DCBF parameter $\gamma$, allowing the safety margin to adjust according to the interaction context.

We adopt the \emph{Adaptive Smooth Tchebycheff Attention} (ASTCH) framework \cite{murillo2026adaptive} to learn the policy by balancing conflicting objectives. ASTCH integrates Smooth Tchebycheff scalarization into an on-policy policy-gradient approach based on Proximal Policy Optimization (PPO). This method, referred to as \emph{Policy-optimization via Adaptive Smooth Tchebycheff Attention} (PASTA)~\cite{murillo2026adaptive}, enables stable learning over non-convex multi-objective trade-offs:
\begin{equation}
\begin{aligned}
\max_{\theta,\phi} 
\mathbb{E}_{o,a \sim \pi_{\theta_k}}
\!\left[
L_t^{\mathrm{CLIP}}(\theta)
- c_1 L_t^{\mathrm{VF}}(\phi)
+ c_2 S[\pi_\theta](o_t)
\right]
\end{aligned}
\end{equation}
\noindent where $\theta$ and $\phi$ represent the parameters of the policy and value networks, respectively, $k$ is the index of the optimization iteration, and $t$ denotes the time step within the sampled environment interaction. $\mathbb{E}(\cdot)$ denotes the expectation taken over the observations visited and actions sampled by the old policy $\pi_{\theta_k}$. $L_t^{\mathrm{CLIP}}(\theta)$ is the multi-objective clipped surrogate advantage utilizing the adaptive smoothness controller, $S[\pi_\theta](o_t)$ is the entropy bonus of the policy evaluated at observation $o_t$ , and $c_1$ and $c_2$ are coefficients weighting the value error and entropy bonus, respectively. More details are provided in \cite{murillo2026adaptive}.

\begin{algorithm}[H]
{\small
\caption{Hierarchical Overtaking Control}
\begin{algorithmic}[1]
\State \textbf{Input:} Ego state ${x}$, opponent states 
\State Transform ego and opponent states to Frenet frame
\State Predict opponent motion over horizon
\If{MIQP update triggered}
    \State Solve MIQP Eqn~(\ref{eq:miqp}) to obtain lateral reference $\{d_k^*\}$
\EndIf
\State Observe ${o}_k$ and compute $\gamma_k \leftarrow \pi_\theta({o}_k)$
\State Solve MPC--CBF Eqn~(\ref{eq:mpc}) using $\{d_k^*\}$ and $\gamma_k$
\end{algorithmic}
\label{algorithm}
}
\end{algorithm}

The policy is trained offline in simulation and deployed online in a
receding-horizon manner. It outputs the barrier decay parameter at each control step $\gamma = \pi_\theta(o)$.
The observation $o_k$ includes the ego vehicle state, relative opponent configuration, track geometry, and the instantaneous safety margin. 
The reward is designed to encourage progress while penalizing safety violations, expressed as
$
r_k = - w_{\mathrm{safe}} \, \mathbb{I}_{\text{col/OOB}} 
      + w_{\mathrm{ot}} \, \mathbb{I}_{\text{overtake}},
$
where $\mathbb{I}$ denotes the indicator function, and 
$w_{\mathrm{safe}} > 0$ and $w_{\mathrm{ot}} > 0$ are scalar coefficients weighting the penalty for collisions or out-of-bounds events and the reward for a successful overtake, respectively. 
The ASTCH-based formulation \cite{murillo2026adaptive} is chosen because the interactive overtaking problem inherently presents a non-convex trade-off space. The discrete, opposing nature of the reward components—where maximizing the probability of a successful overtake strictly conflicts with minimizing the risk of collisions or boundary violations—creates optimization challenges that standard linear scalarization methods may struggle to capture effectively. For comparison, we also evaluated a standard PPO implementation under identical training conditions; however, it exhibited lower success rates and less stable adaptation behavior than the ASTCH-based formulation.

The overall hierarchical overtaking control loop is summarized in Algorithm~\ref{algorithm}.
At each control loop, the ego vehicle receives state estimates of itself and nearby opponents. Opponent states are transformed into the Frenet frame and propagated over a finite horizon using a constant-velocity prediction model. The high-level mixed-integer quadratic program selects the overtaking homotopy and produces a coarse lateral reference trajectory. 
The RL policy then evaluates the observation and determines the CBF decay parameter. The reference trajectory and adapted decay parameter are then passed to the nonlinear MPC augmented with DCBF, which solves the resulting optimal control problem.

\begin{figure*}[h!]
\centering
    \setlength{\fboxsep}{0.1pt}
    \setlength{\fboxrule}{0.2pt} 

    \begin{subfigure}{0.16\textwidth}
        \centering
        \fbox{\includegraphics[width=\linewidth]{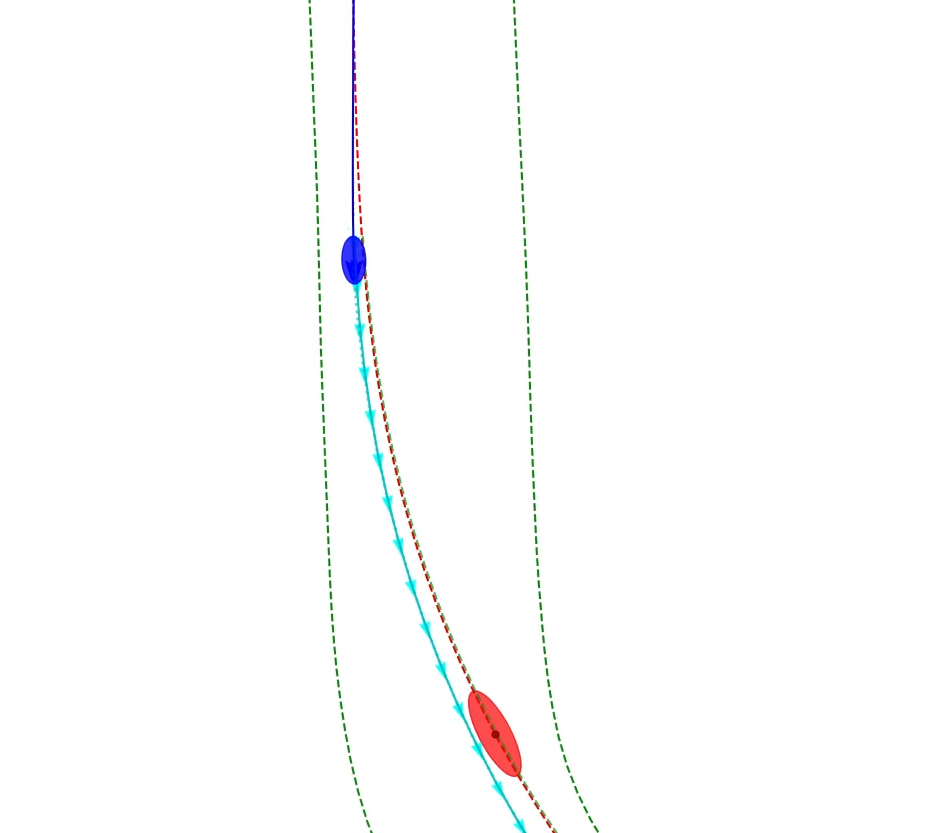}}
        \caption{}
    \end{subfigure}
    \begin{subfigure}{0.16\textwidth}
        \centering
        \fbox{\includegraphics[width=\linewidth]{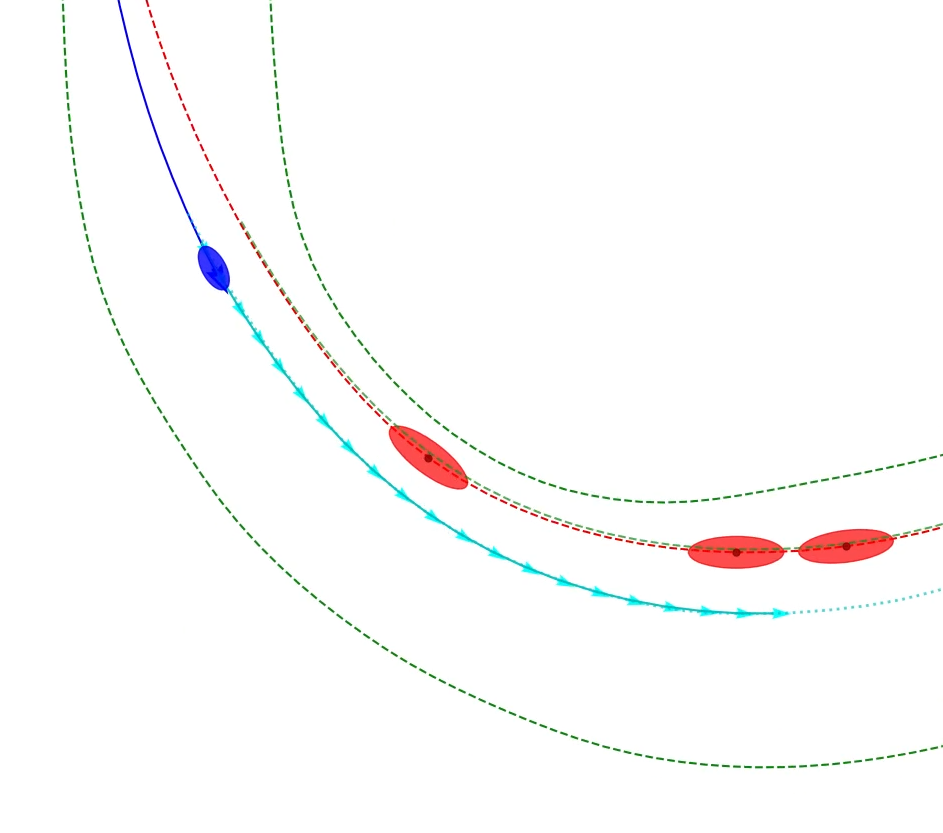}}
        \caption{}
    \end{subfigure}
    \begin{subfigure}{0.16\textwidth}
        \centering
        \fbox{\includegraphics[width=\linewidth]{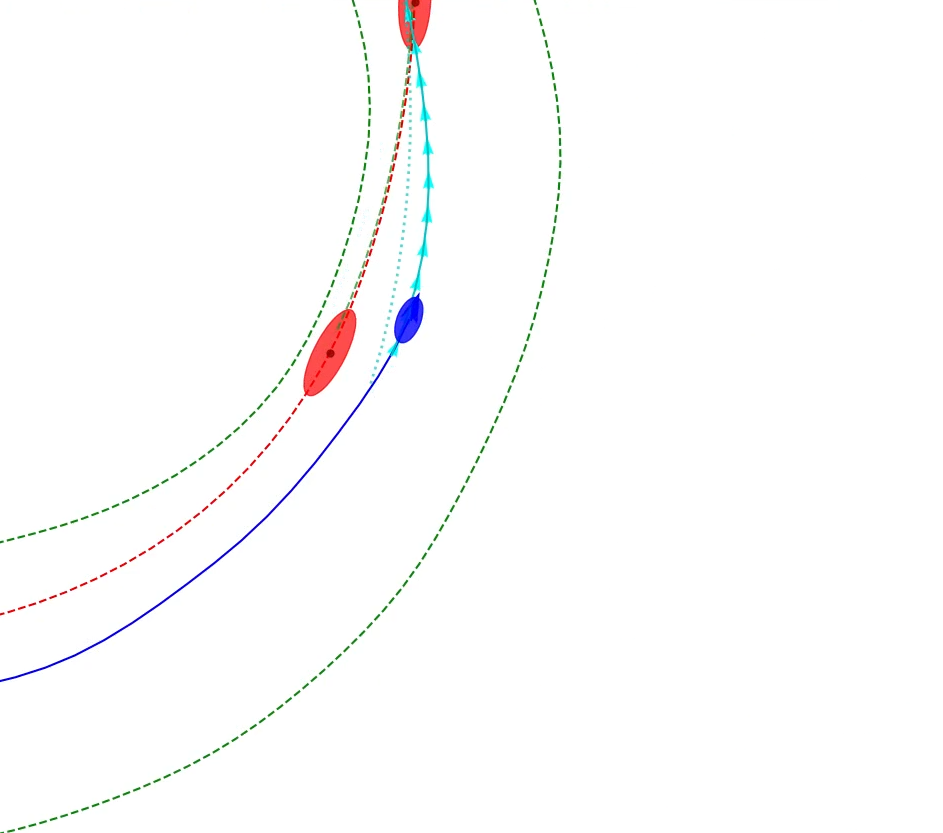}}
        \caption{}
    \end{subfigure}
    \begin{subfigure}{0.16\textwidth}
        \centering
        \fbox{\includegraphics[width=\linewidth]{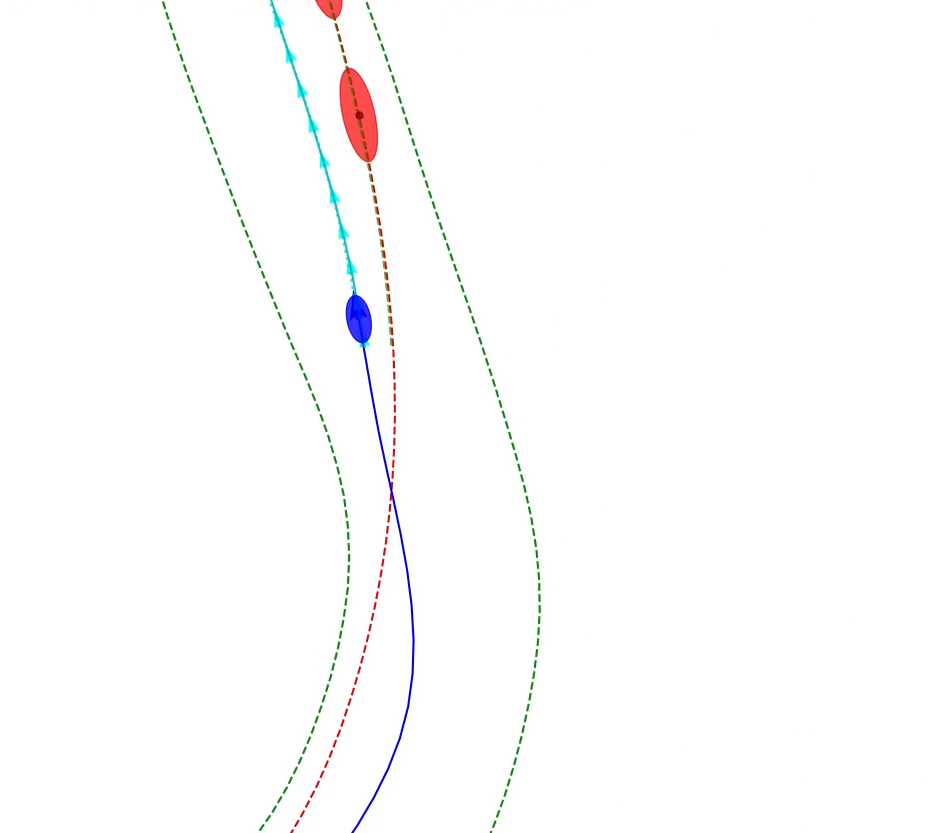}}
        \caption{}
    \end{subfigure}
    \begin{subfigure}{0.16\textwidth}
        \centering
        \fbox{\includegraphics[width=\linewidth]{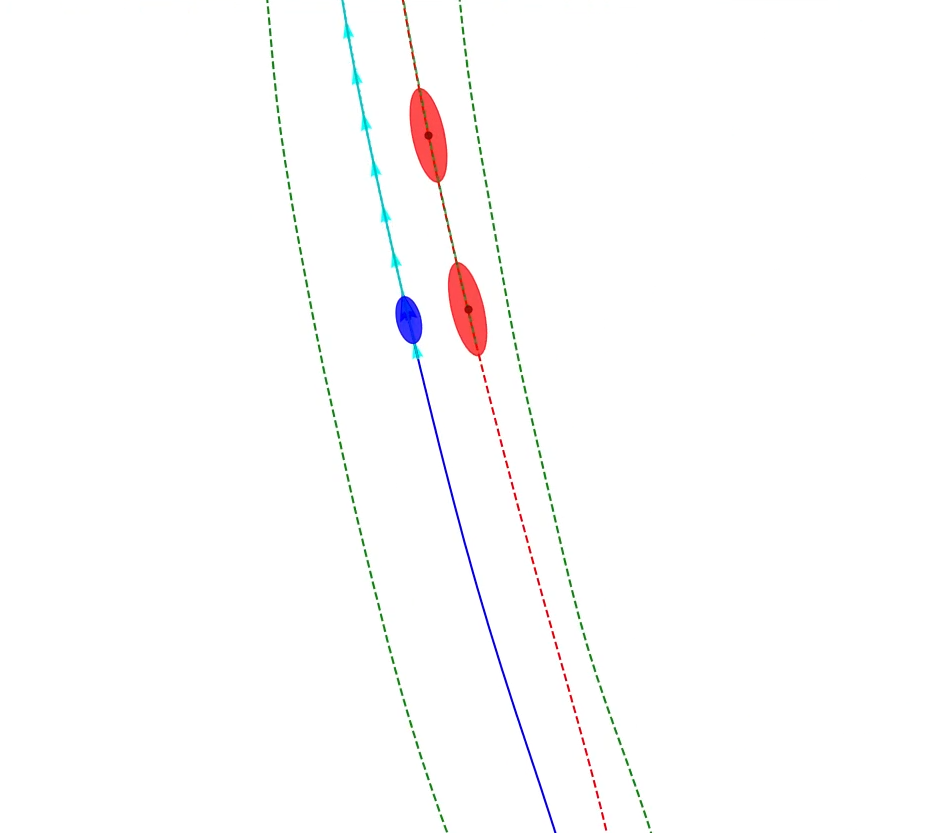}}
        \caption{}
    \end{subfigure}
    \begin{subfigure}{0.16\textwidth}
        \centering
        \fbox{\includegraphics[width=\linewidth]{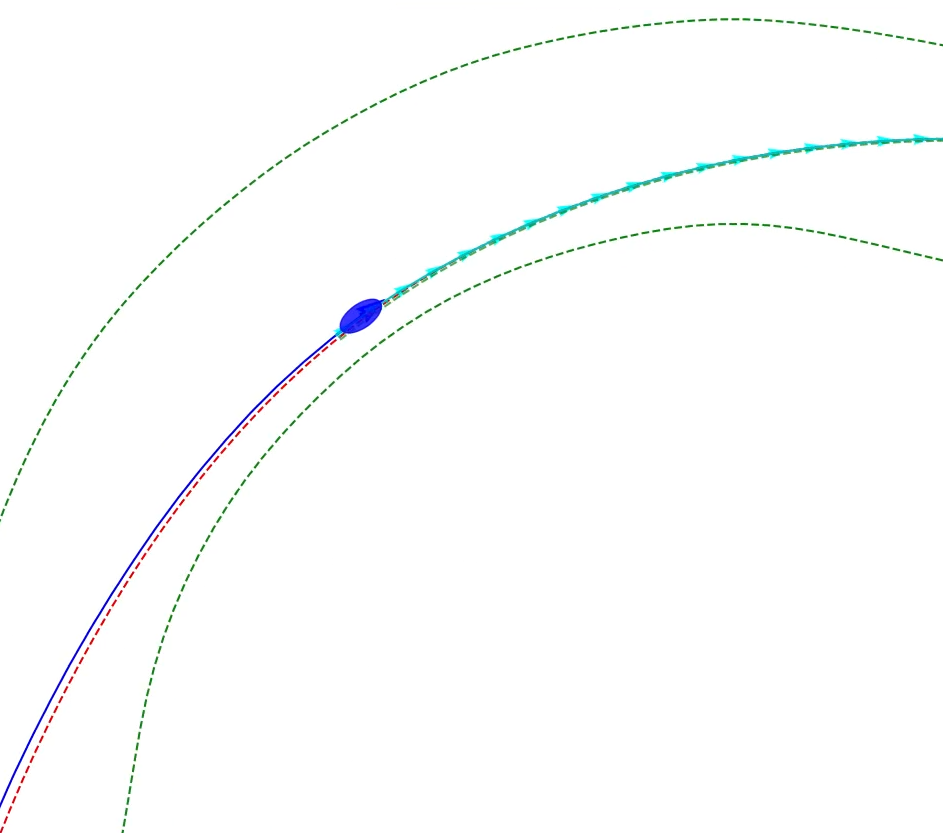}}
        \caption{}
    \end{subfigure}
    
    \caption{\small Sequential snapshots (a--f) of the overtaking maneuver at Turn 2 on the LS track. Additional simulation videos are available at \href{https://github.com/hassanjardali/SOfAR/tree/main/Media}{\textit{github.com/hassanjardali/SOfAR}}.}
    \label{fig:overtake_sequence}
    \vspace{-0.15in}
\end{figure*}

\section{EXPERIMENTS AND RESULTS}

This section details the experimental setup and the simulation environment used to validate our proposed framework, followed by a comparative analysis of the results.

\subsection{Simulation and Training Environment}
A custom Python-based simulation environment was developed to validate the proposed methodology. The simulator employs a nonlinear kinematic bicycle model parameterized to represent the physical characteristics and actuation limits of the Dallara AV-21 autonomous race car. Training and evaluation were conducted across diverse track geometries captured from real-world raceways, including high-speed ovals such as Kentucky Speedway (KS) and Indianapolis Motor Speedway (IMS) and road courses such as Laguna Seca (LS) and Monza. 

For the adaptive (PASTA) algorithm \cite{murillo2026adaptive}, the actor and branched critic networks are parameterized as multi-layer perceptrons with two hidden layers of 256 units each. The inputs to these networks consist of the 12-dimensional continuous state $o_k$ concatenated with the objective preference weights $w$. The continuous action space is bounded to output the decay parameter $\gamma \in [0.05, 1.0]$.

\subsection{Evaluation Metrics}To quantitatively assess the efficacy of the proposed framework in balancing safety and competitiveness, the following primary evaluation metrics are defined:

\begin{itemize}

\item \textbf{Success Rate (\%):}
Percentage of episodes that reach the 200\,s horizon without \textit{collision} or \textit{track-boundary} violation. Episodes terminated early due to safety violations are counted as failures.

\item \textbf{Cumulative Simulation Time (s):}
Total elapsed time in seconds across all episodes until termination. Early termination due to safety violations reduces the accumulated time, making this a robustness measure.

\item \textbf{Laps Completed:}
Total number of laps completed by the ego vehicle across all episodes, reflecting sustained forward progress and average racing speed.

\item \textbf{Overtakes:}
Total number of overtaking events across all episodes. An overtake is detected when the longitudinal distance to an opponent exceeds a predefined threshold, after which the opponent is repositioned ahead.

\end{itemize}

\subsection{Detailed overtaking performance}

Figure~\ref{fig:overtake_sequence} illustrates a representative maneuver at Turn 2 on LS, where the ego vehicle overtakes three opponents aligned on the raceline. The first vehicle is passed on the right; subsequently, the MIQP layer re-solves the discrete decisions and commits to a left-side corridor to overtake the remaining two vehicles. The transition is triggered by MIQP re-optimization, resulting in consistent lateral commitment and smooth closed-loop tracking.

\begin{figure}[t]
    \centering
    \begin{subfigure}{0.48\columnwidth}
        \centering
        \includegraphics[width=\linewidth]{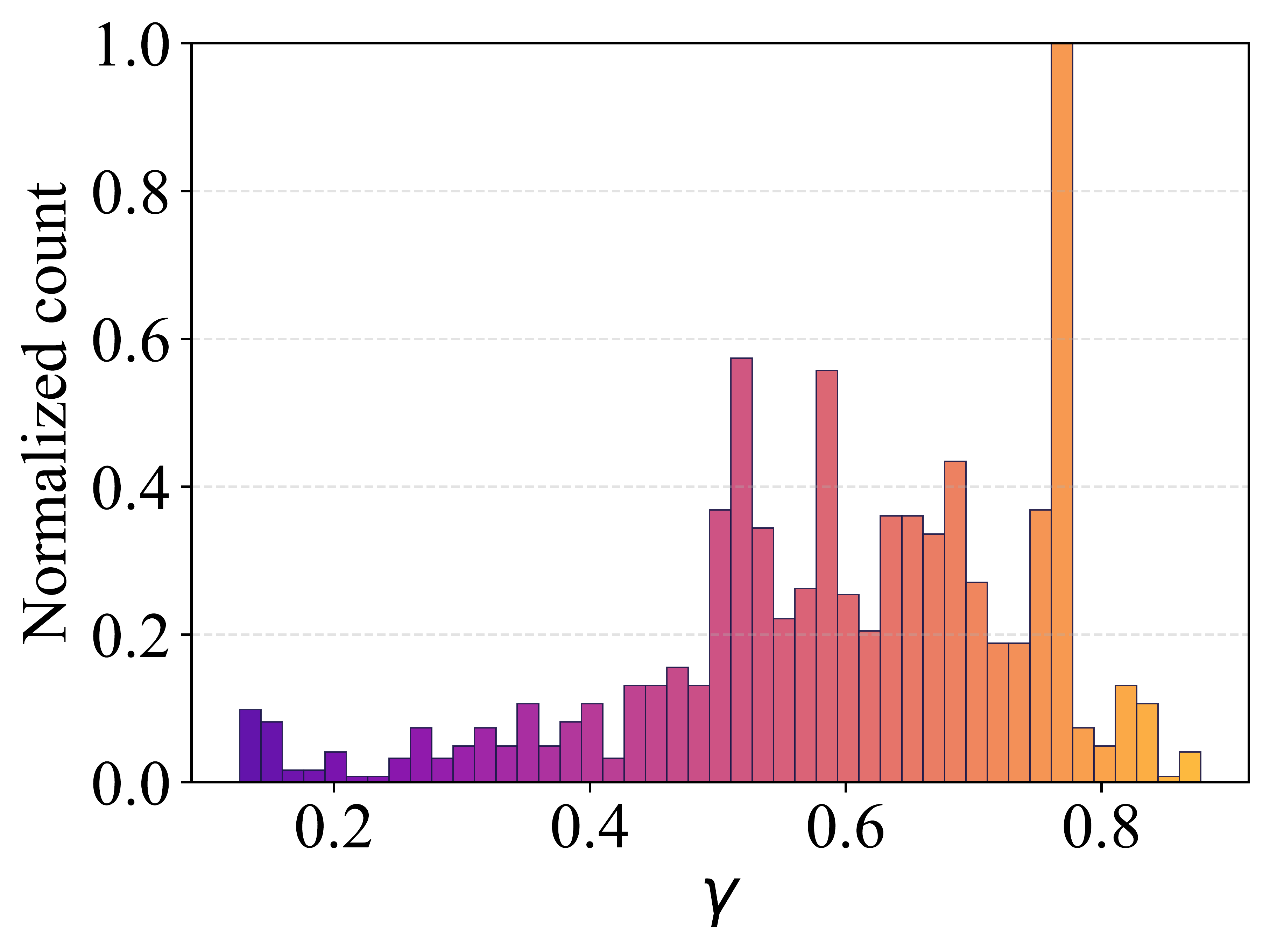}
        \caption{$\gamma$ distribution.}
        \label{fig:gamma_hist}
    \end{subfigure}
    \hfill
    \begin{subfigure}{0.468\columnwidth}
        \centering
        \includegraphics[width=\linewidth]{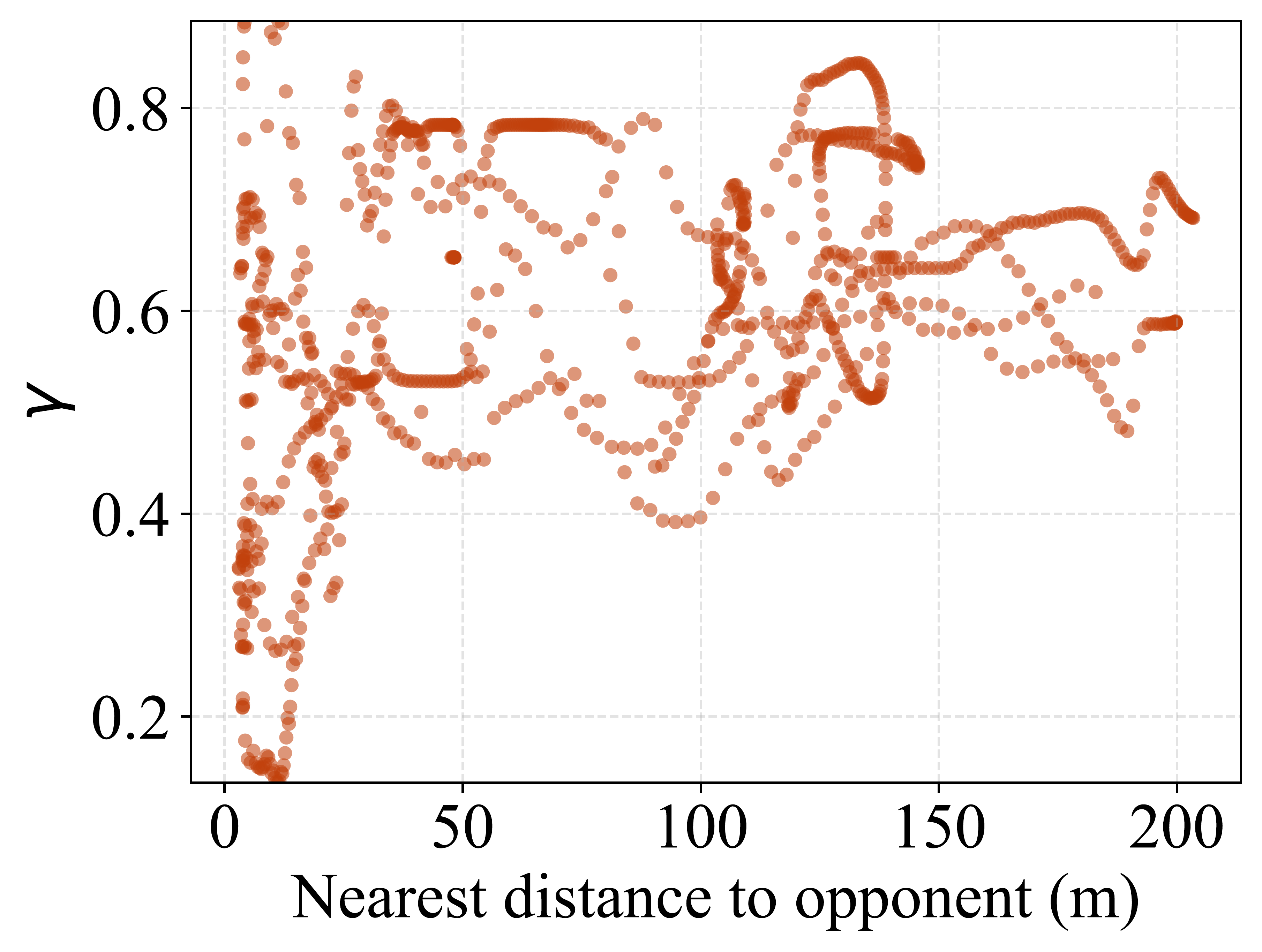}
        \caption{Distance vs. $\gamma$.}
        \label{fig:dist_gamma}
    \end{subfigure}
    \caption{\small Analysis of the policy behavior regarding gamma selection and safety distances.}
    \label{fig:combined_plots}
    \vspace{0.05in}
\end{figure}

The adaptive decay behavior is analyzed in Fig.~\ref{fig:combined_plots}. The histogram in Fig.~\ref{fig:gamma_hist} reveals a pronounced concentration of $\gamma$ values around $0.7$–$0.8$, indicating that the controller predominantly operates in a moderately conservative decay regime during nominal driving. Lower $\gamma$ values occur primarily during close vehicle interactions. Importantly, the distribution shows no persistent saturation at extreme limits, suggesting that the policy does not collapse to boundary values but instead modulates decay within an interior operating range. The distance–$\gamma$ relationship (Fig.~\ref{fig:dist_gamma}) shows that as inter-vehicle distance decreases, $\gamma$ tends to decrease and exhibits increased variance. 
While not strictly monotonic, this trend indicates state-dependent adaptation: as inter-vehicle distance decreases, the policy selects smaller values of $\gamma$, thereby enforcing a more restrictive barrier evolution when safety margins shrink. Conversely, for larger separations, $\gamma$ increases, reflecting less restrictive decay requirements in interaction-free driving conditions.


Together, these observations confirm that overtaking behavior arises from coordinated interaction between discrete homotopy selection (via MIQP), continuous MPC tracking, and smooth, state-dependent decay modulation, rather than from aggressive or discontinuous control actions.

\subsection{Comparative Evaluation}

We compare fixed-decay DCBF controllers with multiple $\gamma \in \{0.05, 0.15, 0.30, 0.50, 0.70, 0.95\}$ against the proposed adaptive decay strategy. All methods are evaluated under identical closed-loop conditions across four tracks (IMS, KS, LS, Monza) and five opponent pace ratios $\rho \in \{0.1, 0.3, 0.5, 0.65, 0.8\}$. Except for the decay parameter selection (fixed or adaptive) and the randomized initial ego state used to generate independent trials, all controller parameters and simulation settings are held constant. Each (track, $\rho$) pair is executed for five episodes, yielding 25 episodes per track and configuration.

Episodes terminate either at the time horizon or upon safety violation (collision or track-boundary violation). Performance is assessed using success rate, cumulative simulation time, completed laps, and overtakes, as defined above. All metrics are aggregated per track to evaluate robustness and competitiveness under varying interaction intensities.


\begin{table}[t]
\centering
\caption{\small Performance comparison across tracks.  
For each metric, \textbf{bold} denotes the best value and \underline{underline} denotes the second-best value within each track.}
\label{tab:full_comparison}
\resizebox{\columnwidth}{!}{
\begin{tabular}{llcccc}
\toprule
\textbf{Track} & \textbf{Strategy} & \textbf{Success Rate [\%]} $\uparrow$ & \textbf{Sim Time [s]} $\uparrow$ & \textbf{Laps} $\uparrow$ & \textbf{Overtakes} $\uparrow$ \\
\midrule

\multirow{7}{*}{IMS}
& Adaptive & \textbf{72.0} & \textbf{4269.1} & \textbf{50.13} & \textbf{796} \\
& $\gamma=0.05$ & 28.0 & 2877.3 & 34.18 & 482 \\
& $\gamma=0.15$ & 44.0 & 3497.8 & 41.63 & 628 \\
& $\gamma=0.30$ & 56.0 & 3697.0 & 43.85 & 619 \\
& $\gamma=0.50$ & 44.0 & 3359.1 & 39.79 & 570 \\
& $\gamma=0.70$ & \underline{68.0} & \underline{4111.5} & \underline{47.86} & \underline{720} \\
& $\gamma=0.95$ & 16.0 & 1934.3 & 18.78 & 411 \\

\midrule

\multirow{7}{*}{KS}
& Adaptive & \underline{48.0} & 3572.6 & 78.06 & 488 \\
& $\gamma=0.05$ & 12.0 & 2144.9 & 46.32 & 203 \\
& $\gamma=0.15$ & 40.0 & 3479.4 & 75.50 & 413 \\
& $\gamma=0.30$ & 44.0 & \textbf{3678.4} & \textbf{80.53} & \textbf{536} \\
& $\gamma=0.50$ & \textbf{56.0} & \underline{3625.7} & \underline{78.81} & \underline{518} \\
& $\gamma=0.70$ & 24.0 & 2996.1 & 64.64 & 436 \\
& $\gamma=0.95$ & 16.0 & 1938.3 & 38.88 & 370 \\

\midrule

\multirow{7}{*}{LS}
& Adaptive & \textbf{76.0} & \underline{4381.5} & \underline{31.54} & \textbf{213} \\
& $\gamma=0.05$ & 0.0 & 1040.3 & 6.00 & 20 \\
& $\gamma=0.15$ & \underline{68.0} & \underline{4206.0} & 28.21 & \underline{171} \\
& $\gamma=0.30$ & 64.0 & \textbf{4639.2} & \textbf{31.68} & 185 \\
& $\gamma=0.50$ & 48.0 & 3657.4 & 27.14 & 162 \\
& $\gamma=0.70$ & 56.0 & 4061.2 & 28.22 & 163 \\
& $\gamma=0.95$ & 12.0 & 1580.2 & 12.00 & 135 \\

\midrule

\multirow{7}{*}{Monza}
& Adaptive & 72.0 & 4300.6 & 25.66 & 136 \\
& $\gamma=0.05$ & 0.0 & 965.8 & 4.88 & 8 \\
& $\gamma=0.15$ & 60.0 & 4173.2 & \underline{26.24} & 125 \\
& $\gamma=0.30$ & 68.0 & 4101.9 & 24.65 & 122 \\
& $\gamma=0.50$ & \textbf{84.0} & \textbf{4482.4} & \textbf{26.80} & \underline{150} \\
& $\gamma=0.70$ & \underline{80.0} & \underline{4472.8} & 25.11 & \textbf{151} \\
& $\gamma=0.95$ & 0.0 & 692.0 & 2.43 & 16 \\

\bottomrule
\end{tabular}
}
\end{table}

Table~\ref{tab:full_comparison} summarizes performance across four tracks. The adaptive policy was trained exclusively on LS, while IMS, KS, and Monza are unseen during training.

\paragraph{Track-Dependent Sensitivity of Fixed $\gamma$}
The optimal fixed decay parameter varies significantly across environments. On IMS, $\gamma=0.70$ achieves the strongest fixed performance (68\% success), whereas LS favors $\gamma=0.15$ (68\%). On Monza, $\gamma=0.50$ yields the highest success rate (84\%), while KS performs best at $\gamma=0.50$ (56\%). No single fixed $\gamma$ simultaneously achieves near-optimal success rates across all tracks. Both small ($\gamma=0.05$) and overly large ($\gamma=0.95$) decay values consistently degrade robustness, confirming strong sensitivity to decay selection.

\paragraph{Generalization Across Tracks}
Despite being trained only on LS, the adaptive controller achieves the highest overall success rate across all tracks (67\% aggregate). It attains 72\% on IMS and 76\% on LS, outperforming all fixed-$\gamma$ configurations on those tracks. While not always matching the single best tuned fixed value on Monza and KS, it maintains competitive success rates (72\% and 48\%, respectively) together with substantial racing activity.
From a cross-track perspective, fixed $\gamma$ values exhibit environment-specific dominance: parameters that perform well on one track may degrade significantly on others. The adaptive controller, by contrast, consistently achieves strong performance without per-track tuning and avoids the catastrophic failures observed for extreme decay selections.

\paragraph{Overall Assessment}
Across all tracks, no single fixed $\gamma$ achieves uniformly strong performance, confirming pronounced environment-specific sensitivity. The adaptive strategy eliminates manual decay selection and attains the highest aggregate success rate (67 out of 100 episodes) while maintaining competitive lap counts and overtaking activity. These results indicate improved robustness to track variation through context-dependent safety modulation.

\begin{table}[h]
\centering
\caption{\small Performance comparison on the LS track with and without MIQP integration.}
\label{tab:miqp_comparison}
\resizebox{\columnwidth}{!}{
\begin{tabular}{llcccc}
\toprule
\textbf{Setting} & \textbf{Strategy} & \textbf{Succ. [\%]} $\uparrow$ & \textbf{Time [s]} $\uparrow$ & \textbf{Laps} $\uparrow$ & \textbf{Overt.} $\uparrow$ \\
\midrule
\multirow{2}{*}{w/o MIQP} 
& Adaptive & 20.0 & 1127.6 & 8.55 & 41 \\
& $\gamma=0.5$     & 10.0 & 1225.9 & 9.81 & 37 \\
\midrule
\multirow{2}{*}{w/ MIQP} 
& Adaptive & \underline{50.0} & \underline{1449.8} & \underline{12.78} & \underline{94} \\
& $\gamma=0.5$     & \textbf{60.0} & \textbf{1633.7} & \textbf{14.59} & \textbf{123} \\
\bottomrule
\end{tabular}
}

\end{table}

\subsection{Effect of MIQP-Based Homotopy Resolution}

Table~\ref{tab:miqp_comparison} evaluates the impact of removing the MIQP overtaking layer on the LS track. In both settings, the MPC and DCBF formulations are identical.

Eliminating the MIQP layer leads to substantial degradation in both safety and racing performance for both strategies. Success rates drop markedly (Adaptive: 50\%→20\%, $\gamma=0.5$: 60\%→10\%), accompanied by reduced simulation time and overtaking activity.
This consistent performance loss across both static and adaptive decay strategies implies that the failure mode is structural rather than parameter-dependent. Without explicit homotopy selection, the downstream continuous MPC is forced to resolve discrete side-selection decisions within a non-convex collision-avoidance region. It frequently traps the solver in local minima, causing oscillatory lateral commitments and eventual safety violations. By enforcing a stable, low-frequency overtaking decision sequence, the MIQP layer successfully improves the closed-loop robustness of the continuous tracking controller. These results confirm that the proposed hierarchical framework is a critical prerequisite for reliable multi-agent autonomous racing. 

\subsection{Computational Performance}

Experiments were conducted on an Intel i7-12700H CPU with 16\,GB RAM.  As shown in Table~\ref{tab:timing_comparison}, the MPC requires $1.60 \pm 0.34$,ms per iteration, the MIQP module incurs $5.49 \pm 1.73$,ms, and the RL policy inference requires $0.82 \pm 0.12$,ms. The overall latency remains below a 20\,ms control cycle (50\,Hz), confirming real-time feasibility. MIQP operates at a lower frequency and does not affect the MPC loop.

\begin{table}[h]
\centering
\small
\caption{\small Computational Latency ($mean \pm std$ in ms).}
\label{tab:timing_comparison}
\begin{tabular}{ccc}
\toprule
\textbf{MPC} & \textbf{MIQP} & \textbf{RL} \\
\midrule
$1.60 \pm 0.34$ & $5.49 \pm 1.73$ & $0.82 \pm 0.12$ \\
\bottomrule
\end{tabular}
\vspace{-0.1in}
\end{table}


\subsection{Hardware Demonstration}

To demonstrate real-world deployability beyond simulation, we conducted hardware experiments (see Fig.~\ref{fig:snapshots}) on an F1TENTH platform ~\cite{o2020f1tenth} using a fixed decay parameter $\gamma$. A virtual opponent was used to reproduce interactive overtaking scenarios, enabling evaluation of the full hierarchical framework—including MIQP planning and MPC–CBF control—under real sensing, actuation, and onboard computation constraints. Due to limited indoor track space and to ensure repeatable testing, experiments were performed with a single physical vehicle. A video of the hardware experiments is available at:
\textit{\href{https://github.com/hassanjardali/SOfAR/tree/main/Media}{github.com/hassanjardali/SOfAR}}.


\section{Conclusion}


This paper presented a hierarchical framework for safe overtaking in multi-vehicle autonomous racing, integrating MIQP-based homotopy resolution, Frenet-frame nonlinear MPC with discrete-time CBF constraints, and an RL policy that adapts the decay parameter $\gamma$ online. Empirical results across four tracks confirm that no single fixed decay performs reliably across environments, whereas adaptive modulation of $\gamma$ achieves the highest aggregate success rate (67\%) without per-track tuning and generalizes beyond its LS training track to unseen configurations. Ablation studies underscore the importance of hierarchical decomposition: removing the MIQP layer substantially degrades performance by forcing the MPC to resolve non-convex side-selection decisions. Computational profiling confirms real-time feasibility at 50\,Hz, and scaled hardware experiments demonstrate deployability under real sensing and actuation constraints.

Several limitations remain. Safety guarantees rely on inactive slack variables in the soft DCBF formulation, and dense interactions may temporarily reduce safety margins. The constant-velocity opponent model and weaker generalization to high-speed oval tracks highlight opportunities for broader training and richer interaction modeling. Finally, the MIQP–MPC decoupling does not ensure global optimality and remains sensitive to baseline parameter selection. Future work will address these aspects toward full-scale autonomous racing deployment.


\section*{Appendix}\label{app:frenet_model}
The bicycle model we applied has the following form
\begin{equation}
\begin{aligned}
\dot{s} &= \frac{v \cos(e + \beta)}{1 - \kappa(s)\, n}, \,
\dot{n} = v \sin(e + \beta), \\
\dot{e} &= \frac{v}{L}\cos\beta \tan\delta - \kappa(s)\dot{s}, \,
\dot{v} = a, \, 
\dot{\delta} = \omega,
\end{aligned}
\label{eq:frenet_kinematic_model}
\end{equation}
where $\kappa(s)$ denotes the curvature of the reference path at longitudinal position $s$, and $L$ represents the vehicle wheelbase, with side-slip angle
$\beta = \arctan\!\left(\frac{l_r}{L}\tan\delta\right)$.
\label{eq:beta}



\section*{Acknowledgment}

This manuscript was refined with the assistance of ChatGPT, which was used to improve clarity, grammar, and overall presentation. 

\addtolength{\textheight}{-12cm} 

\bibliographystyle{IEEEtran}
\bibliography{references} 

@article{mayne2014model,
  title={Model predictive control: Recent developments and future promise},
  author={Mayne, David Q},
  journal={Automatica},
  volume={50},
  number={12},
  pages={2967--2986},
  year={2014},
  publisher={Elsevier}
}

@article{pia2017mixed,
  title={Mixed-integer quadratic programming is in NP},
  author={Pia, Alberto Del and Dey, Santanu S and Molinaro, Marco},
  journal={Mathematical Programming},
  volume={162},
  number={1},
  pages={225--240},
  year={2017},
  publisher={Springer}
}

@article{dollar2021multilane,
  title={Multilane automated driving with optimal control and mixed-integer programming},
  author={Dollar, Robert Austin and Vahidi, Ardalan},
  journal={IEEE Transactions on Control Systems Technology},
  volume={29},
  number={6},
  pages={2561--2574},
  year={2021},
  publisher={IEEE}
}

@inproceedings{miller2018efficient,
  title={Efficient mixed-integer programming for longitudinal and lateral motion planning of autonomous vehicles},
  author={Miller, Christina and Pek, Christian and Althoff, Matthias},
  booktitle={2018 IEEE Intelligent Vehicles Symposium (IV)},
  pages={1954--1961},
  year={2018},
  organization={IEEE}
}

@article{quirynen2024real,
  title={Real-time mixed-integer quadratic programming for vehicle decision-making and motion planning},
  author={Quirynen, Rien and Safaoui, Sleiman and Di Cairano, Stefano},
  journal={IEEE Transactions on Control Systems Technology},
  volume={33},
  number={1},
  pages={77--91},
  year={2024},
  publisher={IEEE}
}

@article{ioan2021mixed,
  title={Mixed-integer programming in motion planning},
  author={Ioan, Daniel and Prodan, Ionela and Olaru, Sorin and Stoican, Florin and Niculescu, Silviu-Iulian},
  journal={Annual Reviews in Control},
  volume={51},
  pages={65--87},
  year={2021},
  publisher={Elsevier}
}

@misc{indyautonomouschallengeIndyAutonomous,
  title = {Indy Autonomous Challenge},
  howpublished = {\url{www.indyautonomouschallenge.com/}}
}

@misc{a2rlAutonomousRace,
  title = {A2RL Autonomous Car Race},
  howpublished = {\url{https://a2rl.io/autonomous-car-race}}
}

@inproceedings{9483029,
  author={Zeng, Jun and Zhang, Bike and Sreenath, Koushil},
  booktitle={Proc. ACC},
  title={Safety-Critical Model Predictive Control with Discrete-Time Control Barrier Function},
  year={2021},
  pages={3882--3889}
}

@inproceedings{8206086,
  author={Buyval, Alexander and Gabdulin, Aidar and Mustafin, Ruslan and Shimchik, Ilya},
  booktitle={Proc. IEEE/RSJ IROS},
  title={Deriving overtaking strategy from nonlinear model predictive control for a race car},
  year={2017},
  pages={2623--2628}
}

@article{Liniger2015AutonomousRacing,
  title={Optimization-Based Autonomous Racing of 1:43 Scale RC Cars},
  author={Liniger, Alexander and Domahidi, Alexander and Morari, Manfred},
  journal={Optim. Control Appl. Meth.},
  volume={36},
  number={5},
  pages={628--647},
  year={2015}
}

@inproceedings{toschi2025modular,
  title={Modular Decision-Making and Drivable Areas for Multi-Agent Autonomous Racing},
  author={Toschi, Alessandro and Prignoli, Francesco and Bertogna, Marko},
  booktitle={Proc. IEEE/RSJ IROS},
  pages={12435--12441},
  year={2025}
}

@inproceedings{stahl2019multilayer,
  title={Multilayer graph-based trajectory planning for race vehicles in dynamic scenarios},
  author={Stahl, Tim and Wischnewski, Alexander and Betz, Johannes and Lienkamp, Markus},
  booktitle={Proc. IEEE ITSC},
  pages={3149--3154},
  year={2019}
}

@inproceedings{werling2010optimal,
  title={Optimal trajectory generation for dynamic street scenarios in a frenet frame},
  author={Werling, Moritz and Ziegler, Julius and Kammel, S{\"o}ren and Thrun, Sebastian},
  booktitle={Proc. IEEE ICRA},
  pages={987--993},
  year={2010}
}

@inproceedings{raji2022motion,
  title={Motion planning and control for multi vehicle autonomous racing at high speeds},
  author={Raji, Ayoub and Liniger, Alexander and Giove, Andrea and Toschi, Alessandro and Musiu, Nicola and Morra, Daniele and Verucchi, Micaela and Caporale, Danilo and Bertogna, Marko},
  booktitle={Proc. IEEE ITSC},
  pages={2775--2782},
  year={2022}
}

@article{gratzer2024two,
  title={Two-layer MPC architecture for efficient mixed-integer-informed obstacle avoidance in real-time},
  author={Gratzer, Alexander L and Broger, Maximilian M and Schirrer, Alexander and Jakubek, Stefan},
  journal={IEEE Trans. Intell. Transp. Syst.},
  volume={25},
  number={10},
  pages={13767--13784},
  year={2024}
}

@article{park2015homotopy,
  author = {J. Park and S. Karumanchi and K. Iagnemma},
  title = {Homotopy-based divide-and-conquer strategy for optimal trajectory planning via mixed-integer programming},
  journal = {IEEE Trans. Robot.},
  volume = {31},
  pages = {1101--1115},
  year = {2015}
}

@inproceedings{9811969,
  author={He, Suiyi and Zeng, Jun and Sreenath, Koushil},
  booktitle={Proc. IEEE ICRA},
  title={Autonomous Racing with Multiple Vehicles using a Parallelized Optimization with Safety Guarantee using Control Barrier Functions},
  year={2022},
  pages={3444--3451}
}

@article{langmann2025reinforcement,
  author = {Langmann, Alexander and Tokarev, Yevhenii and Piccinini, Mattia and Moller, Korbinian and Betz, Johannes},
  title = {Reinforcement learning-based dynamic adaptation for sampling-based motion planning in agile autonomous driving},
  journal = {arXiv:2510.10567},
  year = {2025}
}

@inproceedings{sabouni2024reinforcement,
  title={Reinforcement learning-based receding horizon control using adaptive control barrier functions for safety-critical systems},
  author={Sabouni, Ehsan and Ahmad, HM Sabbir and Giammarino, Vittorio and Cassandras, Christos G and Paschalidis, Ioannis Ch and Li, Wenchao},
  booktitle={Proc. IEEE CDC},
  pages={401--406},
  year={2024}
}

@article{wijayatunga2026learning,
  title={Learning Safety-Guaranteed, Non-Greedy Control Barrier Functions Using Reinforcement Learning},
  author={Wijayatunga, Minduli and Wallace, Nathan and Sukkarieh, Salah and Armellin, Roberto},
  journal={arXiv preprint arXiv:2602.00366},
  year={2026}
}

@inproceedings{10422566,
  author = {G. Jank and M. Rowold and B. Lohmann},
  title = {Hierarchical Time-Optimal Planning for Multi-Vehicle Racing},
  booktitle = {Proc. ITSC},
  pages = {2064--2069},
  year = {2023},
  doi = {10.1109/ITSC57777.2023.10422566}
}

@article{murillo2026adaptive,
  title={Adaptive Smooth Tchebycheff Attention for Multi-Objective Policy Optimization},
  author={Murillo-Gonzalez, Alejandro and Ali, Mahmoud and Liu, Lantao},
  journal={arXiv preprint arXiv:2605.12771. To appear in the Proceedings of Robotics: Science and Systems (RSS)},
  year={2026}
}

@inproceedings{ames2019control,
  author = {Ames, A. D. and Coogan, S. and Egerstedt, M. and Notomista, G. and Sreenath, K. and Tabuada, P.},
  title = {Control Barrier Functions: Theory and Applications},
  booktitle = {Proc. European Control Conf. (ECC)},
  pages = {3420--3431},
  year = {2019}
}

@article{wolsey2007mixed,
  title={Mixed integer programming},
  author={Wolsey, Laurence A},
  journal={Wiley Encyclopedia of Computer Science and Engineering},
  pages={1--10},
  year={2007},
  publisher={Wiley Online Library}
}

@article{o2020f1tenth,
  title={F1tenth: An open-source evaluation environment for continuous control and reinforcement learning},
  author={O'Kelly, Matthew and Zheng, Hongrui and Karthik, Dhruv and Mangharam, Rahul},
  journal={Proceedings of Machine Learning Research},
  volume={123},
  year={2020}
}

\end{document}